%% file: main.tex
\documentclass[runningheads]{llncs}
\usepackage[pagebackref,breaklinks,colorlinks]{hyperref}
\input{macro}

\begin{document}


\title{Automata-Based Steering of Large Language Models for Diverse Structured Generation}
\titlerunning{Automata-Based Steering for Diverse Structured Generation}

\author{Xiaokun~Luan\orcidID{0000-0002-5878-6486}
    \and
    Zeming~Wei\orcidID{0009-0008-2953-0749}
    \and
    Yihao~Zhang\orcidID{0009-0002-0284-1367}
    \and
    Meng~Sun\thanks{Corresponding author.}\orcidID{0000-0001-6550-7396}
}

\authorrunning{X. Luan et al.}

\institute{School of Mathematical Sciences, Peking University, Beijing, China\\
    \email{\{luanxiaokun, sunm\}@pku.edu.cn}\\
    \email{\{weizeming, zhangyihao\}@stu.pku.edu.cn}}

\maketitle

\begin{abstract}
    Large language models (LLMs) are increasingly tasked with generating structured outputs.
    While structured generation methods ensure validity, they often lack output diversity, a critical limitation that we confirm in our preliminary study.
    We propose a novel method to enhance diversity in automaton-based structured generation.
    Our approach utilizes automata traversal history to steer LLMs towards novel structural patterns.
    Evaluations show our method significantly improves structural and content diversity while maintaining comparable generation efficiency.
    Furthermore, we conduct a case study showcasing the effectiveness of our method in generating diverse test cases for testing open-source libraries.

    \keywords{Large language models \and Structured generation \and Diversity}
\end{abstract}

\input{sections/introduction}

\input{sections/background}

\input{sections/study}

\input{sections/method}

\input{sections/evaluation}

\input{sections/related_work}

\input{sections/conclusion}

\begin{credits}
    \subsubsection{\ackname}
    This work was supported by the National Key R\&D Program of China under Grant 2022YFB2702200, National Natural Science Foundation of China (Grant No. 62172019), and Beijing Natural Science Foundation (Grant Nos. QY24035, QY23041).
\end{credits}

\clearpage

\bibliographystyle{splncs04}
\bibliography{references}

\clearpage
\appendix
\input{sections/appendix}

\end{document}

%% file: macro.tex
\usepackage[ruled,vlined]{algorithm2e}
\usepackage{amsmath}
\usepackage{amssymb}
\usepackage{booktabs}
\usepackage{caption}
\usepackage{cite}
\usepackage{cmap}
\usepackage{enumitem}
\usepackage[T1]{fontenc}
\usepackage{graphicx}
\usepackage{hyperref}
\usepackage[utf8]{inputenc}
\usepackage{listings}
\usepackage{mathtools}
\usepackage{microtype}
\usepackage{multirow}
\usepackage{siunitx}
\usepackage{soul}
\usepackage{textcomp}
\usepackage{xcolor}

\newcommand{\widx}[2]{#1\mbox{$[#2]$}}
\newcommand{\wslc}[3]{#1\mbox{$[#2\mathord{:}#3]$}}

\lstdefinestyle{regexstyle}{
    basicstyle=\scriptsize\ttfamily,
    breaklines=true,
    postbreak=\mbox{\textcolor{red}{$\hookrightarrow$}},
    tabsize=2,
    showspaces=false,
    showstringspaces=false,
    showtabs=false,
    captionpos=b,
}

%% file: sections/introduction.tex
\section{Introduction}
\label{sec:introduction}

Large language models (LLMs) have demonstrated remarkable capabilities in a wide range of tasks~\cite{sun2021ernie3,du2022glm,liu2023metamath,wang2024lego}, leading to their rapid adoption across various domains.
As LLMs are increasingly integrated into complex systems, such as AI agents~\cite{guo2024survey,wang2024voyager} and automated code generation~\cite{austin2021program,nijkamp2023codegen}, the demand for reliable and precisely formatted outputs has become critical.
These downstream applications often require LLMs to produce outputs that strictly adhere to predefined structures, such as JSON schemas, API call formats, XML documents, or formal specifications, which are essential for ensuring parsability, correctness, and interoperability in complex systems.
However, standard LLM generation cannot guarantee consistently structured outputs due to issues like hallucination~\cite{ji2023survey}, lack of adherence to instructions~\cite{lou2024survey}, and inconsistency across multiple outputs~\cite{stureborg2024llm}.

Structured generation techniques have emerged to ensure that LLM outputs conform to specified structural requirements~\cite{openai-structured-outputs}.
Prominent examples include tools like Outlines~\cite{willard2023efficient}, SGLang~\cite{zheng2024sglang}, and XGrammar~\cite{dong2024xgrammar}, which leverage finite automata or pushdown automata to guide the token selection process.
Invalid tokens not conforming to the specified grammar are filtered out, ensuring that the generated outputs adhere to the desired structure.
While these approaches have been widely adopted for enforcing output validity, their primary focus has been on correctness.
The diversity of the generated structured outputs, however, often remains overlooked.

The importance of diversity is well-recognized in unconstrained LLM generation~\cite{zhang2021trading,chung2023increasing}, where it fosters creativity and broader coverage of potential responses.
This need is equally pertinent in structured generation downstream applications.
For instance, in software testing, a diverse set of structured test cases (e.g., API call sequences, configuration files) is crucial for achieving high test coverage and uncovering edge cases.
Similarly, AI agents can benefit from diverse structured plans to enhance their adaptability and robustness in complex environments.
However, the diversity of the outputs from the current state-of-the-art structured generation methods has not been thoroughly explored.
Our preliminary study investigates this question by analyzing outputs from the popular structured generation method Outlines.
We observe that the generated samples exhibit limited diversity, tending to repeat common structural patterns.
This finding highlights that existing methods, even when paired with sampling strategies like adjusting temperature, may not effectively explore the full space of valid, diverse structured allowed by the constraints.
This lack of diversity can hinder the performance of downstream applications, as they may rely on a wide range of structured outputs to function effectively.

To bridge this gap, we propose a novel approach to enhance diversity in regular expression constrained generation.
Our core idea is to leverage the history of exploration within the guiding automaton during the generation process.
Specifically, we monitor the states and transitions traversed by the LLM.
This historical information is then used to adaptively adjust the LLM's token selection probabilities, encouraging the model to explore less frequented paths within the automaton.
To maintain generation quality and prevent unproductive exploration, our method incorporates a penalty mechanism to discourage looping in local states and a dynamic range adjustment factor to ensure appropriate guidance intensity.
Our methodology, while targeted at finite automata and regular expressions, is generalizable to pushdown automata and context-free grammars.

Evaluations of our proposed method demonstrate significant improvements in both structural diversity and content diversity compared to baseline method.
These gains are achieved while maintaining a substantial portion (approximately 88\%) of the baseline generation efficiency.
We also conduct ablation studies to confirm the effectiveness of our method under various settings.
Furthermore, the case study on generating test cases for testing open-source libraries illustrates that our method can produce diverse test cases that achieve a higher code coverage compared to those generated by the baseline method.

The main contributions of this paper are as follows:
\begin{enumerate}
    \item A novel method to enhance diversity in automaton-based structured generation, which systematically encourages the LLM to explore less frequented paths within the guiding automaton.
    \item A comprehensive evaluation of our method against the baseline method, demonstrating significant improvements in output diversity while maintaining comparable generation efficiency.
    \item A case study on generating diverse test cases for testing open-source libraries, showcasing the practical applicability of our method in real-world scenarios.
\end{enumerate}

%% file: sections/background.tex
\section{Background}
\label{sec:background}

In this section, we provide a brief overview of large language model generation and structured generation.
Later, we describe the notations and concepts of deterministic finite automata used in our work.

\subsection{Large Language Model Generation}

Large language models generate text by predicting the next token in a sequence following an autoregressive manner.
This process starts with a prompt, and a token is sampled from the model's predicted distribution and appended to the input.
The generation continues until an end-of-sequence token \texttt{EOS} is produced or a maximum length is reached.

Formally, given a vocabulary $\mathcal{V}$ and an input sequence $x = (x_1, x_2, \ldots, x_n) \in \mathcal{V}^n$, the language model $M$ returns a logit vector $z$ for the next token, which is then converted into a probability distribution over the vocabulary $\mathcal{V}$ using the softmax function.
The next token $x_{n+1}$ is sampled from the distribution $\text{softmax}(z)$, defined as follows.
\begin{align}
    z       & = M(x_1, x_2, \ldots, x_n),                                                                                      \\
    x_{n+1} & \sim \text{softmax}(z), \text{ where } \text{softmax}(z)_i = \frac{e^{z_i}}{\sum_{j=1}^{|\mathcal{V}|} e^{z_j}}.
\end{align}
Sampling temperature $T$ controls the randomness of the sampling process by scaling the logits before applying softmax, i.e., $\mathrm{softmax}(z/T)$.
The higher the temperature, the more uniform the distribution becomes, leading to more diverse outputs.
For simplicity, we use $x_{1:n} \coloneq (x_1, x_2, \ldots, x_n)$ to denote the sequence of tokens generated so far.

\subsection{Structured Generation}

Structured generation methods impose constraints on the generation process to ensure adherence to the given grammar.
At each step, tokens that do not conform to the grammar are masked out, and their logits are set to $-\infty$, ensuring that invalid tokens are never sampled.
Formally, let $G$ be a grammar defining a set of valid strings over the vocabulary $\mathcal{V}$.
Given a sequence $x_{1:n} \in \mathcal{V}^n$, the structured generation process can be described as follows.
\begin{align}
    z       & = M(x_{1:n}),                  \\
    z'      & = \text{mask}(x_{1:n}; G) + z, \\
    x_{n+1} & \sim \text{softmax}(z'),
\end{align}
where $\text{mask}(x_{1:n}; G)\subseteq \mathcal{V}$ is a vector of the same size as $z$ consisting of zeros for valid tokens and $-\infty$ for invalid tokens.
Valid tokens $\mathcal{V}_\text{valid}(x_{1:n};G)$ is determined by the grammar $G$ and the current sequence $x_{1:n}$ to obtain the mask, i.e.,
\begin{equation}
    \text{mask}(x_{1:n}; G)_i =
    \begin{cases}
        0,       & \text{if the } i\text{-th token } w_i \in \mathcal{V}_\text{valid}(x_{1:n};G), \\
        -\infty, & \text{otherwise.}
    \end{cases}
\end{equation}
Therefore, the logits of valid tokens are preserved, while the logits of invalid tokens are set to $-\infty$ in $z'$, ensuring that their probabilities are zero.
The grammar $G$ can be defined in various ways, such as regular expressions and context-free grammars.

To efficiently implement the masking process, finite state automata and pushdown automata can be used to represent the grammar and determine the valid tokens at each step.
Specifically, the current state (and the stack) of the automaton is maintained during the generation process.
When a new token is sampled, the automaton transitions to a new state (and updates the stack) based on the sampled token.
The valid tokens for the next step can then be determined by the current state (and stack) of the automaton.
Since the vocabulary is finite, the token-level transition table can be pre-computed to enable efficient masking.

\subsection{Deterministic Finite Automata}

Deterministic finite automata (DFA) are a type of finite state machine that can be used to recognize regular languages.
A DFA $\mathcal{A}$ is defined as a 5-tuple $(Q, \Sigma, \delta, q_0, F)$, where $Q$ is a finite set of states, $\Sigma$ is the alphabet (set of input symbols), $\delta: Q \times \Sigma \to  Q$ is the transition function, which maps a state and an input symbol to the next state, $q_0 \in Q$ is the initial state, and $F \subseteq Q$ is the set of accepting states.
The language defined by the DFA is the set of strings that lead to an accepting state when processed by the automaton, i.e., $L(\mathcal{A}) = \{ w \in \Sigma^* \mid \delta^*(q_0, w) \in F \}$, where $\delta^*$ is the extended transition function inductively defined as $\delta^*(q, \epsilon) = q$ and $\delta^*(q, wa) = \delta(\delta^*(q, w), a)$ with $\epsilon$ being the empty string.

Given a state $q$ and a string $w$, we denote the sequence of states visited when processing $w$ in state $q$ as $\mathrm{States}(q, w) = (q_1, \ldots, q_m)$ with $q_i = \delta^*(q, \wslc{w}{}{i})$, where $\wslc{w}{}{i}$ is the prefix\footnote{Square brackets (e.g., $w_i[j]$ and $w[i:j]$) are for string indexing (0-based) and slicing.} of $w$ of length $i$.
Similarly, we denote the sequence of transitions taken during the processing of $w$ as $\mathrm{Transitions}(q, w) = (t_1, t_2, \ldots, t_m)$, where $t_i = (q_{i-1}, \widx{w}{i}, q_i)$ is the transition from state $q_{i-1}$ to state $q_i$ on input symbol $\widx{w}{i}$.
We assume that the alphabet $\Sigma$ consists of all input symbols that make up tokens in the vocabulary $\mathcal{V}$, i.e., $\mathcal{V}\subseteq\Sigma^*$.
In the following, we will use $w_i$ to denote some string, e.g., a token in the vocabulary $\mathcal{V}$, and $w_i$ with square brackets $\widx{w_i}{j}$ to denote the $j$-th symbol in $w_i$.

To efficiently guide the LLM generation process, one needs to first pre-compute the subset of valid tokens for each state of the DFA and construct a token transition table.
For this purpose, we first introduce the concepts of \emph{live states} and \emph{dead states}.
A state $q$ is considered \emph{live} if there exists a path from $q$ to an accepting state, and it is \emph{dead} if there is no such path.
We denote the set of live states as $Q_\text{live}$ and the set of dead states as $Q_\text{dead}$.
The set of valid tokens that can be generated from a given state $q$ is defined as the subset of the vocabulary that can lead to a live state, i.e., $\mathcal{V}_\text{valid}(q) = \{ w \in \mathcal{V} \mid \delta^*(q, w) \in Q_\text{live} \}$.
Based on this, we can construct a token transition table $T$, where each entry $T(q, w)$ contains the next state after processing the token $w$ from state $q$, i.e., $T(q, w) = \delta^*(q, w)$.

%% file: sections/study.tex
\section{How Diverse is Structured Generation?}
\label{sec:study}

\subsection{A Preliminary Study}

Although structured generation methods guarantee that the generated outputs adhere to the given grammar, not all desirable outputs in the grammar can be generated in practice.
This is due to the fact that the sampling process is stochastic and influenced by the model's prediction, which in turn relies on the training data that may impose biases.

In this section, we conduct a preliminary study to investigate the diversity of structured generation.
We select representative, non-trivial regular expressions as constraints, and generate 1000 samples for each grammar using a state-of-the-art structured generation method.
To evaluate the diversity of the generated samples, we measure finite automata coverage and count Distinct N-grams.
We also briefly examine the effect of varying sampling temperature on output diversity.
In the rest of this section, we present the details of our experimental setup and findings.

\subsection{Sample Generation}

Two regular expressions are selected for our study: $G_\text{email}$~\cite{rfc5322} and $G_\text{color}$~\cite{css-color-regex}, which are designed to capture \emph{all} valid email addresses and CSS color codes, respectively.
These grammars are not as simple as they may seem, as they also cover many edge cases, such as using IPv4 in email addresses and various color formats in CSS.
The minimal DFA for each grammar has 43 and 1309 states, and 1594 and 7495 transitions, respectively.
The complete regular expressions are provided in Appendix~\ref{app:regex}.

We use Qwen2.5-1.5B-Instruct, a \SI{1.5}{B} parameter LLM, to generate 1000 samples for each constraint, following the structured generation method proposed by Willard et al.~\cite{willard2023efficient}.
This method, implemented in the popular Python library \texttt{outlines}~\cite{outlines} (over 11k GitHub stars at the time of writing), has proven to be effective for such tasks.
The rationale of this method has been presented in Section~\ref{sec:background}.
We adopt the default multinomial sampling strategy with a temperature of 1.0 and 1.5, representing the normal and high sampling temperatures, respectively.
All samples are generated independently with a maximum length of 18 tokens, which is sufficient to cover most valid cases.

\subsection{Diversity Evaluation}

We evaluate the diversity of the generated samples from two perspectives: structural diversity and content diversity.
Regarding structural diversity, we measure the coverage of DFA states and transitions by the generated samples.
Specifically, we compile the regular expression into a minimal DFA and then count the number of unique states and transitions that are visited by these samples.
Intuitively, the more diverse the samples are, the more structural patterns they will cover, leading to a higher coverage of states and transitions.
On the contrary, a low coverage indicates that some states and transitions are never reached by the given samples, suggesting that the model is not fully exploring the grammar.
Formally, given a minimal DFA $\mathcal{A} = (Q, \Sigma, \delta, q_0, F)$ and a set of samples $S=\{s_i\}_{i=1}^n$ generated by the model, we introduce the following metrics to measure the coverage of the DFA states and transitions:
\begin{align}
    \mathrm{StateCov}(S) & = \frac{\lvert\bigcup_{i=1}^{n} \mathrm{States}(q_0, s_i)\rvert}{\lvert Q \rvert},                                                              \\
    \mathrm{TransCov}(S) & = \frac{\lvert\bigcup_{i=1}^{n} \mathrm{Transitions}(q_0, s_i)\rvert}{\lvert\{ (q,a,q')\in Q\times\Sigma\times Q \mid \delta(q,a)=q' \}\rvert},
\end{align}
where $\mathrm{States}(q_0,s_i)$ and $\mathrm{Transitions}(q_0,s_i)$ are the sequences of states and transitions traversed by the minimal DFA $\mathcal{A}$ when processing the sample $s_i$.

To evaluate content diversity, we use the commonly used Distinct N-grams metric~\cite{li2016diversity}, which counts the number of unique n-grams in the samples.
Formally, given a set of samples $S$, the Distinct N-grams metric is defined as follows:
\begin{equation}
    \mathrm{Distinct-}n(S) = \lvert \{ \wslc{s}{i}{i+n} \mid s \in S, \lvert s\rvert \geq n, 0\leq i \leq |s| - n \} \rvert.
\end{equation}
This metric captures the local sequence variety of the generated samples.
The greater the $\mathrm{Distinct-}n$ value, the more diverse the samples are.

\subsection{Results and Findings}

\begin{table}[t]
    \centering
    \caption{Diversity evaluation results of preliminary study.}
    \label{tab:preliminary-study}
    \begin{tabular}{lS[table-format=5.2]S[table-format=5.2]S[table-format=4.2]S[table-format=4.2]}
        \toprule
        \multirow{2}{*}{Metric} & \multicolumn{2}{c}{$G_\text{email}$}           & \multicolumn{2}{c}{$G_\text{color}$}                                                                                               \\
        \cmidrule(lr){2-3}       \cmidrule(lr){4-5}
                                & {Temp. 1.0}                                    & {Temp. 1.5}                                    & {Temp. 1.0}                             & {Temp. 1.5}                             \\
        \midrule
        \textrm{StateCov} (\%)  & 18.60                                          & 23.26                                          & 16.96                                   & 31.55                                   \\
        \textrm{TransCov} (\%)  & 20.45                                          & 20.83                                          & 7.59                                    & 12.26                                   \\
        Average length          & 53.6                                           & 68.4                                           & 15.3                                    & 12.8                                    \\
        \textrm{Distinct-}2     & \multicolumn{1}{l}{\hspace{17.4pt}1610}        & \multicolumn{1}{l}{\hspace{17.4pt}1558}        & \multicolumn{1}{l}{\hspace{14.3pt}458}  & \multicolumn{1}{l}{\hspace{14.3pt}635}  \\
        \textrm{Distinct-}3     & \multicolumn{1}{l}{\hspace{17.4pt}\num{10172}} & \multicolumn{1}{l}{\hspace{17.4pt}\num{12526}} & \multicolumn{1}{l}{\hspace{14.3pt}1295} & \multicolumn{1}{l}{\hspace{14.3pt}1991} \\
        \bottomrule
    \end{tabular}
\end{table}

The main results of our preliminary study are reported in Table~\ref{tab:preliminary-study}, where we additionally report average length of the generated samples.
Clearly, the finite automata coverage values for both grammars are relatively low, indicating that the model has not fully explored the DFA even though it has generated 1000 samples.
Regarding the Distinct N-grams metric, the $\mathrm{Distinct-}2$ and $\mathrm{Distinct-}3$ values seem satisfactory.
But when considering the average length of the generated samples, the Distinct N-grams values are not as impressive as they appear, suggesting that there are many repeated patterns in the generated samples.
For example, there are about \num{53600} bigrams in the generated email addresses, but only about 3\% of them are unique.

Theoretically, increasing the sampling temperature can make the sampling process more random, which helps the model explore more edge cases.
However, as shown in Table~\ref{tab:preliminary-study}, the DFA coverage of states and transitions have not improved significantly when we increase the sampling temperature to 1.5.
Most edge cases in the grammars are still not covered, despite that we have used a relatively high temperature.
For example, the regular expression $G_\text{email}$ allows using double quotes in the local part of the email address (e.g., \texttt{"user"@example.com}), but none of the generated samples contain this case.
Similarly, the regular expression $G_\text{color}$ allows using HSL and HSLA color formats (e.g., \texttt{hsl(120, 100\%, 50\%)}) and the \texttt{lch} and \texttt{lab} functions, which are also absent in the generated samples.

The results of our preliminary study indicate that the state-of-the-art structured generation method is not fully exploring the grammar, leading to a lack of diversity in the generated samples.
Increasing the sampling temperature cannot significantly improve the diversity of the generated samples.
This is due to the model's tendency to follow its prediction based on natural language modeling rather than fully considering the grammar constraints.
Such a lack of diversity may significantly limit the applicability of structured generation methods in scenarios where diverse outputs are desired.

%% file: sections/method.tex
\section{Method}
\label{sec:method}

\subsection{Overview}

\begin{algorithm}[t]
    \caption{Diversity-enhanced Structured Generation}
    \label{alg:diversity}
    \KwIn{The number of samples $n$, a minimal DFA $\mathcal{A} = (Q, \Sigma, \delta, q_0, F)$ of the given regex, a token transition table $T$, a hyperparameter $\gamma$}
    \KwOut{A set of generated samples $S=\{s_i\}_{i=1}^{n}$}
    $S = \{\}$\;
    Initialize global transition counter $C$ and local state counter $C_\text{loc}$\;
    \For{$i=1$ \KwTo $n$}{
    $q \gets q_0$\;
    $s_i \gets \epsilon$\;
    Reset local state counter $C_\text{loc}$\;
    \While{True}{
        $z\gets M(prompt + s_i)$\;
        Compute the $\mathrm{range}(q,z)$ and $\mathrm{adjust}(q)$\;
        $z'_j = z_j + \mathrm{mask}(q)_j + \gamma \cdot \mathrm{range}(q,z)\;\mathrm{adjust}(q)_j$ for all $j$\;
        Sample a token $w$ from the distribution $\mathrm{Softmax}(z')$\;
        \If{$w$ \normalfont is \texttt{EOS}}{\textbf{break}\;}
        $s_i\gets s_i + w$\;
        $C_\text{loc}(q_{j+1}) \gets C_\text{loc}(q_{j+1}) + 1$ for each $q_j\in\mathrm{States}(q,w)$\;
        $q \gets T(q,w)$\;
    }
    $C(q_j,q_{j+1})\gets C(q_j,q_{j+1}) + 1$ for each $q_j\in\mathrm{States}(q_0,s_i)$\;
    $S\gets S \cup \{ s_i \}$\;
    }
    \KwRet{$S$}
\end{algorithm}

The lack of diversity in structured generation arises from the model's non-awareness of the legal paths under the grammar constraints, especially those that are rare in the natural language context.
To improve the diversity of structured generation, we encourage the model to explore different paths in the automata by adjusting the logits based on history transitions.
This is achieved by keeping track of the transitions made by the model during the entire generation process to identify valid paths in the grammar that are rarely explored.
In addition, we penalize the tokens that lead to frequently visited states to avoid looping into local optima.
The adjustment terms are adaptively scaled based on the range of the logits to avoid over-penalizing or over-rewarding the tokens.

Algorithm~\ref{alg:diversity} summarizes the overall process of our method, where the main difference between our method and the standard structured generation method is that we adjust the logits based on history transitions.
The adjustment is achieved by maintaining two counters during the generation process.
In the following, we describe the main components of our adjustment terms in detail.

\subsection{Encouraging Exploration}

Suppose a minimal DFA $\mathcal{A} = (Q, \Sigma, \delta, q_0, F)$ of the given regular expression and its corresponding token transition table $T$ are constructed.
By design, the minimal DFA ensures that ambiguities and overlapping transitions are resolved.
Recall that a token transition $T(q, w)$ makes several transitions to the next state $\delta^*(q, w)$ after generating a new token $w$, going through a sequence of states $\mathrm{States}(q, w)$.
If there are two adjacent states $q_i$ and $q_{i+1}$ in this sequence and there have never been any transition moving from $q_i$ to $q_{i+1}$ during the whole generation process, then we can encourage the model to sample the token $w$ in state $q$ so that a new path is explored.
Intuitively, the less such a state pair appears in the whole generation process, the more we should reward the model to sample it, thus potentially increasing the structural diversity of the generated samples.

To achieve this, we maintain a global transition counter $C$ to keep track of the number of times each state bigram (i.e., state pair) has been taken during the whole generation process.
Formally, $C$ is a mapping from a state pair to the number of times it has been taken, i.e., $C: Q\times Q \to \mathbb{N}$.
$C(q,q')$ is initialized to zero at the beginning of the generation process.
This counter is updated after a valid sample $s$ is generated\footnote{Invalid samples are possible due to reaching maximum tokens before generating the \texttt{EOS} token.}, where $C(q_i,q_{i+1})$ is incremented by one for each state pair $(q_i,q_{i+1})$ in the sequence $\mathrm{States}(q_0,s)$.
With such a global counter, we can encourage the model to explore infrequently visited paths by adjusting the logits of the next token.
Specifically, assume the current state is $q$ and the valid tokens are $w_1, w_2, \ldots, w_k\in\mathcal{V}_\text{valid}(q)$.
For each token transition $T(q, w_i)$, we measure if it is worth exploring by quantifying the least frequently visited state pair in its transition sequence $\mathrm{States}(q,w_i)$, defined as follows:
\begin{equation}
    E(q,w_i) = \min_{q_j\in \mathrm{States}(q,w_i)} C(q_j,q_{j+1}).
\end{equation}
Intuitively, $E(q,w_i)$ represents how many times the least frequently visited state bigram has been taken in the transition sequence of $w_i$ from state $q$.
To encourage the model to explore such less frequently visited paths, we add a reward term $\mathrm{reward}(q)$ to the logits of the next token, defined as follows:
\begin{equation}
    \mathrm{reward}(q)_i = \begin{cases}
        \dfrac{\log \bigl(1 + \sum\limits_{j=1}^{k}E(q,w_j)\bigr)}{1 + E(q,w_i)}, & \text{if } w_i\in\mathcal{V}_\text{valid}(q),     \\
        0,                                                                        & \text{if } w_i\not\in\mathcal{V}_\text{valid}(q),
    \end{cases}
\end{equation}
where $k$ is the number of valid tokens in state $q$.
The reward term $\mathrm{reward}(q)$ is added to the logits of the next token $x_{n+1}$, so that as the model generates more samples, the reward term will favor under-explored paths and thus increasing the diversity of the generated samples.

\subsection{Avoiding Looping into Local Optima}

However, as the global transition counter $C$ is only updated after a valid sample is generated, it may result in a situation where the model repeatedly predicts a token that leads back to the same state.
For example, after generating a few samples, the reward term may favor certain token $w_i$ in state $q$ that leads to the same state $q$ again, thus causing the model to loop into local optima and fail to reach accepting states.

To avoid this, we introduce a local state counter $C_\text{loc}: Q\to \mathbb{N}$ to keep track of the number of times each state has been visited during the generation process of a single sample.
Therefore, $C_\text{loc}$ is similar to $C$, except that it is reset to zero at the beginning of each sample generation and updated after a new token is sampled.
We similarly introduce $m(q,w_i)$ to denote the maximum number of times a state has been visited in the state sequence of a valid token $w_i$ from state $q$, defined as follows:
\begin{equation}
    m(q,w_i) = \max_{q_j\in \mathrm{States}(q,w_i)} C_\text{loc}(q_j).
\end{equation}
Based on this, we employ a penalty term $\mathrm{penalty}(q)$ to adjust the logits of the next token, defined as follows:
\begin{equation}
    \mathrm{penalty}(q)_i = \begin{cases}
        \beta (1 + m(q,w_i)), & \text{if } w_i\in\mathcal{V}_\text{valid}(q),     \\
        1,                    & \text{if } w_i\not\in\mathcal{V}_\text{valid}(q),
    \end{cases}
\end{equation}
where $\beta$ is a hyperparameter that controls the intensity of the penalty.
The reward term $\mathrm{reward}(q)_i$ is divided by the penalty term $\mathrm{penalty}(q)_i$ before being added to the logits, i.e.,
\begin{equation}
    \mathrm{adjust}(q)_i = \frac{\mathrm{reward}(q)_i}{\mathrm{penalty}(q)_i}.
\end{equation}
Therefore, the penalty term will prevent the model from repeatedly predicting the same token that leads to the same state, even if the reward term favors it.

\subsection{Adaptive Scaling}

To better adapt the reward and penalty terms to the varying scale of the logits, we introduce an adaptive scaling method to adjust the reward and penalty terms based on the range of the logits and a hyperparameter $\gamma$, i.e.,
\begin{equation}
    z'_i = z_i + \mathrm{mask}(q)_i + \gamma \cdot \mathrm{range}(q,z)\;\mathrm{adjust}(q)_i,
\end{equation}
where $z_i$ is the original logits of the token $w_i$, $\mathrm{mask}(q)_i$ is the mask term for the token $w_i$, and $\mathrm{range}(q,z)$ is the range of the logits in state $q$, defined as the difference between the maximum and minimum logits of valid tokens in state $q$:
\begin{equation}
    \mathrm{range}(q,z) = \max_{w_i\in\mathcal{V}_\text{valid}(q)} z_i - \min_{w_i\in\mathcal{V}_\text{valid}(q)} z_i.
\end{equation}
This adaptive scaling method allows the model to adjust the reward and penalty terms based on the scale of the logits, thus avoiding over-penalizing or over-rewarding the tokens.
Meanwhile, the hyperparameter $\gamma$ controls the strength of the adjustment, serving as a lever to balance exploration and exploitation.
A higher $\gamma$ value amplifies the reward for less-frequented paths, encouraging the model to generate more diverse outputs.
Conversely, a lower $\gamma$ value leads the model to rely more on its original high-confidence predictions, effectively refining and improving upon the quality of more probable outputs.

%% file: sections/evaluation.tex
\section{Evaluation}
\label{sec:evalution}

This section evaluates the proposed method for enhancing diversity in structured generation.
We compare its performance with a state-of-the-art baseline, Outlines, focusing on diversity and generation efficiency.
Ablation studies are conducted to further assess the method's performance under high-temperature settings and evaluate the contribution of each component.
Finally, we present a case study showcasing the practical application of our method in generating diverse test cases for testing open-source libraries.
Our implementation is publicly available\footnote{\url{https://github.com/luan-xiaokun/diverse-structured-generation}}.

\subsection{Experimental Setup}

\subsubsection{Grammar and Language Model}

We conduct our experiments on four regular expressions, including $G_\text{email}$, $G_\text{color}$, $G_\text{json}$, and $G_\text{bomb}$.
The first two, $G_\text{email}$ and $G_\text{color}$, were introduced in our preliminary study in Section~\ref{sec:study}.
They represent common real-world formats with intricate standardized rules and many edge cases.
The third grammar, $G_\text{json}$, describes a simple JSON object with three fields (i.e., name, gender, age) and represents a common application scenario in structured generation.
Finally, $G_\text{bomb}$ serves as a negative constraint, allowing any string except those containing the word ``bomb'' (case-insensitive).
Such negative constraints are essential for applications like content moderation and safety filtering.
The complete regular expressions are provided in Appendix~\ref{app:regex}.

We use Qwen2.5-1.5B-Instruct, a \SI{1.5}{B} instruction fine-tuned large language model, for structured generation in our experiments.
Unless otherwise specified, we adopt a default multinomial sampling strategy with a temperature of 1.0.
The max length of the generated samples is set to 18 tokens, except for $G_\text{json}$, which is set to 54 tokens.
This setting is sufficient for the grammars we are using.
The prompts used for the generation are as follows:
\begin{itemize}
    \item $G_\text{email}$: Give me an email address.
    \item $G_\text{color}$: Give me a CSS color code.
    \item $G_\text{json}$: Give me a JSON object, which has three fields: name (a string), gender (male or female), age (an integer).
    \item $G_\text{bomb}$: Tell me how to make a bomb.
\end{itemize}

\subsubsection{Baseline and Hyperparameters}

We compare our method with the state-of-the-art baseline method Outlines.
The hyperparameters $\beta$ and $\gamma$ of our method are set to 3 and 0.5, respectively.
Both methods use the same default multinomial sampling strategy with a temperature of 1.0 unless otherwise specified.
For each experiment, we generate 1000 samples per grammar using both our method and the baseline method on a platform with an NVIDIA GeForce RTX 3060Ti GPU and an AMD Ryzen\;9 7900X CPU with \SI{32}{GB} memory.

\subsubsection{Evaluation Metrics}

We employ two sets of metrics to evaluate structural and content diversity.
For structural diversity, we measure the DFA state coverage $\mathrm{StateCov}$, transition coverage $\mathrm{TransCov}$, and  path coverage $\mathrm{PathCov}$ of the generated samples, where the path coverage is defined as the ratio of the number of unique state bigrams covered by the generated samples to the total number of state bigrams in the DFA.
For content diversity, we use the Distinct N-grams and the Vendi score~\cite{friedman2023vendi}, a commonly used metric for evaluating the diversity of datasets.
The Vendi score requires a semi-definite similarity function over the samples, which is a similarity function for strings in our case.
We adopt the weighted-degree kernel with shifts kernel~\cite{raetsch2005rase} as the string similarity function, which considers the number of common substrings between two strings.

\subsection{Performance Comparison}

\begin{table}[t]
    \centering
    \caption{Structural diversity evaluation results.}
    \label{tab:structural}
    \begin{tabular}{llS[table-format=2.2]S[table-format=2.2]S[table-format=2.2]S[table-format=2.2]S[table-format=2.2]S[table-format=2.2]}
        \toprule
        \multirow{2}{*}{Grammar} & \multirow{2}{*}{DFA Size} & \multicolumn{2}{c}{\textrm{StateCov} (\%)} & \multicolumn{2}{c}{\textrm{TransCov} (\%)} & \multicolumn{2}{c}{\textrm{PathCov} (\%)}                                                  \\
        \cmidrule(lr){3-4} \cmidrule(lr){5-6} \cmidrule(lr){7-8}
                                 &                           & {Baseline}                                 & {Ours}                                     & {Baseline}                                & {Ours}          & {Baseline} & {Ours}          \\
        \midrule
        $G_\text{email}$         & 43\,/\,1594               & 18.60                                      & \bfseries 95.35                            & 20.45                                     & \bfseries 31.56 & 13.68      & \bfseries 77.78 \\
        $G_\text{color}$         & 1309\,/\,7495             & 16.96                                      & \bfseries 62.49                            & 7.59                                      & \bfseries 24.94 & 9.12       & \bfseries 42.05 \\
        $G_\text{json}$          & 216\,/\,\num{10192}       & 31.94                                      & \bfseries 56.48                            & 2.04                                      & \bfseries 6.66  & 12.60      & \bfseries 33.11 \\
        $G_\text{bomb}$          & 12\,/\,1213               & 50.00                                      & \bfseries 83.33                            & 12.12                                     & \bfseries 28.69 & 27.45      & \bfseries 70.59 \\
        \bottomrule
    \end{tabular}
\end{table}

\begin{table}[t]
    \centering
    \caption{Content diversity evaluation results.}
    \label{tab:content}
    \begin{tabular}{lS[table-format=2.1]S[table-format=2.1]S[table-format=4]S[table-format=4]S[table-format=5]S[table-format=5]S[table-format=3.1]S[table-format=3.1]}
        \toprule
        \multirow{2}{*}{Grammar} & \multicolumn{2}{c}{Average Length} & \multicolumn{2}{c}{\textrm{Distinct-}2} & \multicolumn{2}{c}{\textrm{Distinct-}3} & \multicolumn{2}{c}{Vendi Score}                                                               \\
        \cmidrule(lr){2-3} \cmidrule(lr){4-5} \cmidrule(lr){6-7} \cmidrule(lr){8-9}
                                 & {Baseline}                         & {Ours}                                  & {Baseline}                              & {Ours}                          & {Baseline} & {Ours}          & {Baseline} & {Ours}          \\
        \midrule
        $G_\text{email}$         & 53.6                               & 51.8                                    & 1610                                    & \bfseries 1670                  & 10172      & \bfseries 10333 & 702.1      & \bfseries 707.7 \\
        $G_\text{color}$         & 15.3                               & 11.1                                    & 458                                     & \bfseries 679                   & 1295       & \bfseries 2039  & 98.0       & \bfseries 169.9 \\
        $G_\text{json}$          & 56.8                               & 53.6                                    & 354                                     & \bfseries 1639                  & 657        & \bfseries 2881  & 14.7       & \bfseries 56.3  \\
        $G_\text{bomb}$          & 82.6                               & 79.5                                    & 1080                                    & \bfseries 2025                  & 4335       & \bfseries 5655  & 477.0      & \bfseries 491.5 \\
        \bottomrule
    \end{tabular}
\end{table}

\subsubsection{Diversity Enhancement}

We first evaluate the performance of our method and the baseline method in terms of structural and content diversity.
The results are shown in Table~\ref{tab:structural} and Table~\ref{tab:content}, respectively, where the DFA size $n/m$ indicates that the minimal DFA has $n$ states and $m$ transitions.
We use bold font to highlight the best results between the two methods for each grammar.
Our method significantly outperforms the baseline method in terms of both structural and content diversity across all four grammars.
The DFA state coverage, transition coverage, path coverage, and the Vendi score of our method are improved by 45\%, 12\%, 40\%, and 90\% on average compared to the baseline method.
The average length of the generated samples is slightly lower than that of the baseline method, which is expected since our method encourages the model to explore more diverse paths.

As the minimal DFA has a dead state dedicated to handle invalid inputs, achieving 100\% state coverage is impossible.
Taking this into account, our method has covered all live states in the DFA of $G_\text{bomb}$.
Another figure worth noting is that the transition table of $G_\text{json}$'s DFA are very dense, with \num{10192} transitions in total.
This is due to the regex of $G_\text{json}$ has a dot-star pattern, which allows any character to appear in the string.
Consequently, its transition coverage is relatively low compared to the other three grammars.

\subsubsection{Efficiency Analysis}

\begin{table}[t]
    \centering
    \caption{Tokens generated per second of our method and the baseline method.}
    \label{tab:efficiency}
    \begin{tabular}{lS[table-format=2.2]S[table-format=2.2]S[table-format=2.2]}
        \toprule
        Grammar          & {Baseline (TPS)} & {Ours (TPS)} & {Percentage}         \\
        \midrule
        $G_\text{email}$ & \bfseries 33.44  & 32.80        & \SI{98.09}{\percent} \\
        $G_\text{color}$ & \bfseries 34.39  & 31.49        & \SI{91.57}{\percent} \\
        $G_\text{json}$  & \bfseries 23.64  & 18.16        & \SI{76.82}{\percent} \\
        $G_\text{bomb}$  & \bfseries 25.69  & 22.60        & \SI{87.97}{\percent} \\
        \bottomrule
    \end{tabular}
\end{table}

The proposed method has a higher computational overhead than the baseline method due to the additional logits adjustment step.
We analyze the efficiency of our method by measuring the number of tokens generated per second (TPS) and compare it with the baseline method.
Table~\ref{tab:efficiency} shows the TPS of both methods for each grammar and the percentage of TPS of our method compared to the baseline method.
The results show that our method is slower than the baseline method, with an average TPS of 88.8\% of the baseline method.
This is mainly caused by the computation of the reward term and the penalty term.
However, considering the gain in diversity, the trade-off is acceptable.

\subsection{Ablation Studies}

\subsubsection{High-Temperature Generation}

\begin{table}[t]
    \centering
    \caption{Structural diversity evaluation results under temperature 1.5.}
    \label{tab:ablation-structural}
    \begin{tabular}{lS[table-format=2.2]S[table-format=2.2]S[table-format=2.2]S[table-format=2.2]S[table-format=2.2]S[table-format=2.2]}
        \toprule
        \multirow{2}{*}{Grammar} & \multicolumn{2}{c}{\textrm{StateCov} (\%)} & \multicolumn{2}{c}{\textrm{TransCov} (\%)} & \multicolumn{2}{c}{\textrm{PathCov} (\%)}                                                  \\
        \cmidrule(lr){2-3} \cmidrule(lr){4-5} \cmidrule(lr){6-7}
                                 & {Baseline}                                 & {Ours}                                     & {Baseline}                                & {Ours}          & {Baseline} & {Ours}          \\
        \midrule
        $G_\text{email}$         & 23.26                                      & \bfseries 90.70                            & 20.83                                     & \bfseries 32.56 & 17.95      & \bfseries 76.92 \\
        $G_\text{color}$         & 31.55                                      & \bfseries 61.65                            & 12.26                                     & \bfseries 24.18 & 16.78      & \bfseries 40.68 \\
        $G_\text{json}$          & 33.33                                      & \bfseries 56.48                            & 3.87                                      & \bfseries 6.76  & 14.34      & \bfseries 32.04 \\
        $G_\text{bomb}$          & 75.00                                      & \bfseries 83.33                            & 30.26                                     & \bfseries 35.94 & 50.98      & \bfseries 74.51 \\
        \bottomrule
    \end{tabular}
\end{table}
\begin{table}[t]
    \centering
    \caption{Content diversity evaluation results under temperature 1.5.}
    \label{tab:ablation-content}
    \begin{tabular}{lS[table-format=2.1]S[table-format=2.1]S[table-format=4]S[table-format=4]S[table-format=5]S[table-format=5]S[table-format=3.1]S[table-format=3.1]}
        \toprule
        \multirow{2}{*}{Grammar} & \multicolumn{2}{c}{Average Length} & \multicolumn{2}{c}{Distinct-2} & \multicolumn{2}{c}{Distinct-3} & \multicolumn{2}{c}{Vendi Score}                                                               \\
        \cmidrule(lr){2-3} \cmidrule(lr){4-5} \cmidrule(lr){6-7} \cmidrule(lr){8-9}
                                 & {Baseline}                         & {Ours}                         & {Baseline}                     & {Ours}                          & {Baseline} & {Ours}          & {Baseline} & {Ours}          \\
        \midrule
        $G_\text{email}$         & 68.4                               & 63.6                           & 1558                           & \bfseries 1814                  & 12526      & \bfseries 12585 & 792.0      & \bfseries 802.8 \\
        $G_\text{color}$         & 12.8                               & 11.3                           & 635                            & \bfseries 720                   & 1991       & \bfseries 2166  & 143.1      & \bfseries 158.4 \\
        $G_\text{json}$          & 59.1                               & 61.7                           & 1132                           & \bfseries 2604                  & 3127       & \bfseries 5570  & 65.4       & \bfseries 121.8 \\
        $G_\text{bomb}$          & 89.6                               & 85.8                           & 5154                           & \bfseries 6805                  & 15375      & \bfseries 17812 & 819.8      & \bfseries 841.5 \\
        \bottomrule
    \end{tabular}
\end{table}

\begin{table}[t]
    \centering
    \caption{Perplexity of generated samples constrained by $G_\text{bomb}$.}
    \label{tab:perplexity-bomb}
    \begin{tabular}{SSS}
        \toprule
        {Temperature} & {Baseline (PPL)} & {Ours (PPL)} \\
        \midrule
        1.0           & 54.6             & 138.0        \\
        1.5           & 55133.3          & 81710.8      \\
        \bottomrule
    \end{tabular}
\end{table}

To evaluate the performance of our method under high-temperature settings, we re-generate the samples with a temperature of 1.5.
The results are shown in Table~\ref{tab:ablation-structural} and Table~\ref{tab:ablation-content}.
As expected, the diversity of samples generated by baseline method improves compared with the default temperature setting.
However, our method still outperforms the baseline method in terms of both structural and content diversity.
Interestingly, all the metrics of our method slightly decrease compared to the default temperature setting.
This occurs because our method and temperature scaling represent two distinct, competing forces for promoting diversity.
Our method directly manipulates logits to widen their range, while temperature scaling smooths the final probability distribution.
When $T>1$, this smoothing effect partially counteracts our explicit logit adjustments, leading to the observed slight decrease in diversity.
Therefore, the slight decrease in the diversity of our method compared with the default temperature setting is expected.

We further analyze the perplexity of the generated sampled constrained by $G_\text{bomb}$ to assess the quality of the generated samples, since $G_\text{bomb}$ is the only grammar that allows natural language generation among the four grammars.
The perplexity is calculated using another larger model, Phi4-mini-instruct, which is a \SI{4}{B} instruction fine-tuned LLM.
Intuitively, a lower perplexity indicates that the generated texts are more natural and fluent.
The results are shown in Table~\ref{tab:perplexity-bomb}, where the perplexity of the generated samples under temperature 1.5 is significantly higher than that under temperature 1.0 for both methods.
Such significant quality drop is not acceptable for natural language generation.
As a result, although the baseline method could achieve a higher diversity with a higher temperature, the generated samples are neither of high quality nor diverse.
On the other hand, our method achieves a good balance between diversity and quality under the default temperature setting.

\subsubsection{Component Analysis}

\begin{table}[t]
    \centering
    \caption{Ablation study on the effectiveness of each component.}
    \label{tab:component}
    \begin{tabular}{lS[table-format=2.2]S[table-format=2.2]S[table-format=2.2]S[table-format=3]S[table-format=4]S[table-format=3.1]}
        \toprule
        \multirow{2}{*}{Components} & \multicolumn{3}{c}{DFA Coverage (\%)} & {\multirow{2}{*}{\textrm{Distinct-}2}} & {\multirow{2}{*}{\textrm{Distinct-}3}} & {\multirow{2}{*}{Vendi Score}}                 \\
        \cmidrule(lr){2-4}
                                    & \textrm{StateCov}                     & \textrm{TransCov}                      & \textrm{PathCov}                       &                                &       &       \\
        \midrule
        All                         & 62.49                                 & 24.94                                  & 42.05                                  & 679                            & 2039  & 169.9 \\
        No $\mathrm{reward}(q)$     & 15.30                                 & 7.20                                   & 8.20                                   & 463                            & 1213  & 93.7  \\
        No $\mathrm{penalty(q)}$    & {---}                                 & {---}                                  & {---}                                  & {---}                          & {---} & {---} \\
        No $\mathrm{range}(q,z)$    & 23.50                                 & 9.40                                   & 12.70                                  & 461                            & 1278  & 96.8  \\
        \bottomrule
    \end{tabular}
\end{table}

We conduct an ablation study to analyze the effectiveness of each core component of our method, including the reward term $\mathrm{reward}(q)$, the penalty term $\mathrm{penalty}(q)$, and the logits range adjustment factor $\mathrm{range}(q,z)$.
We remove one component at a time and evaluate the performance of the modified method on the $G_\text{color}$ grammar.
Table~\ref{tab:component} shows the results of the ablation study.

We first remove the reward term by setting $\mathrm{reward}(q)_i=1$ for all $i$.
As expected, the performance drops significantly, even a bit lower than the baseline method.
This demonstrates that the reward term is essential for our method to enhance diversity.
After removing the penalty term by setting $\mathrm{penalty}(q)_i=1$ for all $i$, we observe that the generation process becomes unstable and that the model struggles to generate valid samples.
As more samples are generated, the model tends to repeat certain patterns (e.g., ``oklch(18.27777...'') that quickly reach the max token limit, resulting in a very low success rate on generating valid samples.
Therefore, we conclude that the penalty term is crucial for the stability of our method, as it avoids the model from getting stuck in a local optimum.
We further remove the logits range adjustment factor.
The structural diversity metrics are slightly better than the baseline method, and the content diversity metrics are on par with the baseline method.
This indicates that without the adaptive scaling provided by the logits range factor, the strength of the adjustment is insufficient to significantly improve diversity.
In summary, all three components contribute to the overall performance of our method.

\subsection{Case Study: Test Case Generation}

\begin{table}[t]
    \centering
    \caption{Branch coverage of generated test cases.}
    \label{tab:case-study}
    \begin{tabular}{llS[table-format=3]S[table-format=2.2]S[table-format=2.2]}
        \toprule
        \multirow{2}{*}{Grammar} & \multirow{2}{*}{Library}  & {\multirow{2}{*}{LoC}} & \multicolumn{2}{c}{Branch Coverage (\%)}                   \\
        \cmidrule(lr){4-5}
                                 &                           &                        & {Baseline}                               & {Ours}          \\
        \midrule
        $G_\text{email}$         & \texttt{email\_validator} & 792                    & 46.19                                    & \bfseries 59.08 \\
        $G_\text{color}$         & \texttt{webcolors}        & 441                    & 78.04                                    & \bfseries 83.18 \\
        \bottomrule
    \end{tabular}
\end{table}

To evaluate the effectiveness of our method in downstream tasks, we conduct a case study on using LLMs to generate test cases for testing open source third-party libraries.
We select two popular Python libraries, \texttt{email\_validator}~\cite{email-validator} and \texttt{webcolors}~\cite{webcolors}, which are mainly used for validating email addresses and converting CSS color codes, respectively.

We utilize the generated samples of $G_\text{email}$ and $G_\text{color}$ generated by baseline and our method to test these libraries.
The branch coverage of the generated test cases is reported in Table~\ref{tab:case-study}.
The results show that our method achieves a higher branch coverage than the baseline method with 12.89\% and 5.14\% improvement on \texttt{email\_validator} and \texttt{webcolors}, demonstrating its effectiveness in generating diverse test cases.
As most generated samples are valid, they cannot trigger part of the library's logic intended to handle invalid inputs.
On the other hand, some samples adhere to the given regular expressions but are not valid in the default context of the library, which contributes to the branch coverage.
For example, the \texttt{email\_validator} library does not accept IPv4 addresses in the email field by default\footnote{Such feature is only supported when a corresponding option is turned on.}, but such variations are valid according to the regex.

To summarize, the proposed diversity-enhanced structured generation method can generate more diverse test cases than the baseline method, which is beneficial for improving the branch coverage of the generated test cases.

%% file: sections/related_work.tex
\section{Related Work}
\label{sec:related-work}

\subsection{Structured Generation of Large Language Models}

Structured generation is proposed to improve the quality of LLM-generated outputs by enforcing constraints during the generation process.
Various methods have been developed to facilitate structured generation, including Guidance\cite{guidance}, Outlines~\cite{willard2023efficient}, XGrammar~\cite{dong2024xgrammar}, and SGLang~\cite{zheng2024sglang}.
Among them, Guidance employs a template-based approach, where users define templates with placeholders and the model fills in the placeholders with generated tokens.
In contrast, Outlines proposes a finite state machine-based approach to mask invalid tokens based on the current automaton state, and it supports both regular expressions and context-free grammars (CFGs).
XGrammar focuses on building an engine for efficient CFG-constrained generation through token mask caching and other various optimization techniques.
SGLang functions as a domain-specific language embedded in Python, featuring a rich front end for defining structured tasks and a highly optimized back end for efficient execution.
Except for Guidance's template system, these prominent methods are fundamentally based on finite automata or pushdown automata.
Their primary distinctions often lie in the specific optimization strategies for their generation and the expressiveness of the primitives they offer to users.
Consequently, while these methods differ in usability and generation efficiency, they have generally not prioritized the diversity of the generated structured outputs, often yielding a limited range of structural variations.
Our proposed method is orthogonal to these existing engines and front ends, allowing for easy integration into them to enhance their output diversity.

\subsection{Diversity Enhancement in Text Generation}

Diversity is widely recognized as a critical attribute of high-quality text generation.
Before the widespread adoption of LLMs, the importance of diversity was already acknowledged, and various methods were proposed to promote it.
For instance, Shi et al.~\cite{shi2018toward} employed inverse reinforcement learning to learn a reward function that encourages diverse outputs.
Xu et al.~\cite{xu2018diversity} proposed Diversity-Promoting Generative Adversarial Networks that assign lower reward for repetitive text and higher reward for novel generations.
Following the advent of LLMs, increasing sampling temperature has been a common heuristic to boost output diversity, though often at the expense of coherence or factual accuracy~\cite{chung2023increasing}.
The trade-off between diversity and other quality aspects is a recurring theme in text generation research.
For example, Shao et al.~\cite{shao2021controllable} introduced a controllable text generative model based on Conditional Variational Autoencoders to manage this balance via a tunable hyperparameter.
Zhao et al.~\cite{zhao2023loft} presented a novel method called LoFT to enhance diversity while maintaining faithfulness in logical table-to-text generation.
To generate long and diverse text outputs from structured data, Shao et al.~\cite{shao2019long} proposed a Planning-based Hierarchical Variational Model, which segments input data into a sequence of groups and generates sentences for each group.
Our approach is complementary to these methods, specifically targeting the enhancement of diversity in automaton-based structured generation.

%% file: sections/conclusion.tex
\section{Conclusion and Future Work}
\label{sec:conclusion}

In this paper, we identified the limitation of state-of-the-art structured generation methods in producing diverse outputs.
To enhance diversity, we proposed a novel method that encourages LLMs to explore new paths in the automata by leveraging the history of traversed states and transitions.
Our evaluations demonstrated that our method significantly improves both structural and content diversity while maintaining comparable generation efficiency.
We also conducted ablation studies to show that simply increasing sampling temperature in standard structured generation is insufficient for achieving substantial diversity improvements and could degrade output quality.
Our case study on generating structured test cases for software testing further illustrated the practical benefits of our approach.
Our findings highlight the importance of exploring the full space of valid structured outputs and suggest that our method can be a valuable addition to existing structured generation techniques.

For future work, we plan to generalize our method to more expressive Context-Free Grammars (CFGs) and develop adaptive strategies for the exploration-exploitation trade-off.
Furthermore, integrating with parser combinators could enable application to general-purpose programming languages, enhancing tasks like automated code generation.

%% file: sections/appendix.tex
\appendix

\section*{Appendix}
\label{app:regex}

\renewcommand{\thelstlisting}{\arabic{lstlisting}}
\setcounter{lstlisting}{0}

The regular expressions $G_\text{email}$, $G_\text{json}$, and $G_\text{bomb}$ used in our study are provided in the following listings.
$G_\text{color}$ is exceptionally complex as it covers various color formats.
Therefore, we provide a GitHub repository link\footnote{\url{https://github.com/Kyza/color-regex/}} that contains the complete $G_\text{color}$ expression instead of including it here.

\begin{lstlisting}[caption={The regular expression $G_\text{email}$.}, label={lst:email-regex}]
(?:[a-z0-9!#$%&'*+/=?^_`{|}~-]+(?:\.[a-z0-9!#$%&'*+/=?^_`{|}~-]+)*|\"(?:[\x01
-\x08\x0b\x0c\x0e-\x1f\x21\x23-\x5b\x5d-\x7f]|\\[\x01-\x09\x0b\x0c\x0e-\x7f])
*\")@(?:(?:[a-z0-9](?:[a-z0-9-]*[a-z0-9])?\.)+[a-z0-9](?:[a-z0-9-]*[a-z0-9])?
|\[(?:(?:(2(5[0-5]|[0-4][0-9])|1[0-9][0-9]|[1-9]?[0-9]))\.){3}(?:(2(5[0-5]|[0
-4][0-9])|1[0-9][0-9]|[1-9]?[0-9])|[a-z0-9-]*[a-z0-9]:(?:[\x01-\x08\x0b\x0c\x
0e-\x1f\x21-\x5a\x53-\x7f]|\\[\x01-\x09\x0b\x0c\x0e-\x7f])+)\])$
\end{lstlisting}

\begin{lstlisting}[caption={The regular expression $G_\text{json}$.}, label={lst:json-regex}]
\{\s*\"name\":\s*\"(?:.+?)\",\s*\"gender\":\s*\"(?:fe)?male\",\s*\"age\":\s*
\d+\s*\}$
\end{lstlisting}

\begin{lstlisting}[caption={The regular expression $G_\text{bomb}$.}, label={lst:bomb-regex}]
(?:[^bB]|[bB][^oO]|[bB][oO][^mM]|[bB][oO][mM][^bB])+
\end{lstlisting}